\documentstyle[twocolumn,epsf]{ETRW}

\begin{document}

\title{Robust Parsing of Spoken Dialogue Using Contextual Knowledge and
  Recognition Probabilities} 

\author{{\bf Gerhard Hanrieder, G\"unther G\"orz}\\[5mm]
Bavarian Research Center for Knowledge Based Systems (FORWISS)\\
Am Weichselgarten 7, 91058 Erlangen, Germany
}

\maketitle

\newcommand{\bra}{\left[ \begin{array}{l}}
\newcommand{\ket}{\end{array} \right]}
\newcommand{\angbra}{\left< \begin{array}{l}}
\newcommand{\angket}{\end{array} \right>}

\section*{ABSTRACT}
In this paper we describe the linguistic processor of a spoken
dialogue system. The parser receives a word graph from the recognition
module as its input. Its task is to find the best path through the graph.
If no complete solution can be found, a robust mechanism for selecting
multiple partial results is applied. We show how the information content
rate of the results can be improved if the selection is based on an
integrated quality score combining word recognition scores and
context-dependent semantic predictions. Results of parsing word
graphs with and without predictions are reported.

\section{INTRODUCTION}
The linguistic processing (LP) component of a spoken
dialogue system (SDS) must be robust in order to deal with recognition
errors and spontaneous speech phenomena. 

\begin{figure}[htb]
\epsfxsize=8cm
\centerline{\epsffile{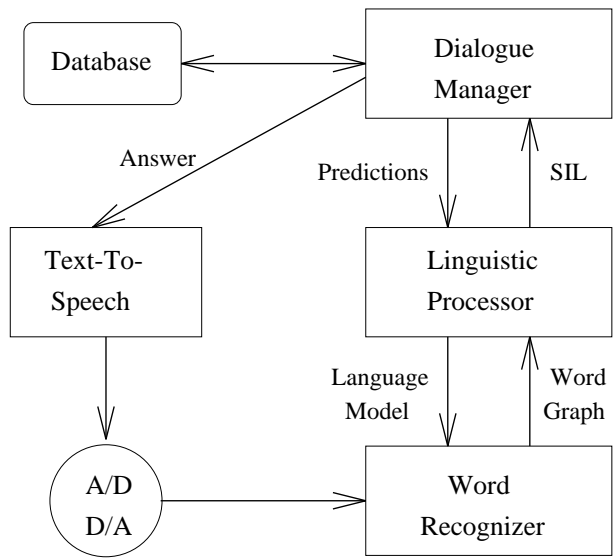}}
\label{archi}
\caption{System architecture}
\end{figure}

In the following we describe our approach towards robustness. This LP was
developed in the project SYSLID ({\bf SY}ntactic and {\bf S}emantic {\bf
  LI}nguistic Processing for Spoken {\bf D}ialogue Systems). It is
fully integrated into the Daimler-Benz SDS [3] for German train
timetable inquiries. The architecture of this system is shown in Figure
\ref{archi}. 

It has been pointed out in [5] that a robust parser
which may deliver multiple partial results has to cope with the problem of
deciding which partial results should be selected. The solution suggested in
[5] relies on the assignment of a quality score to each
partial solution generated during parsing by means of a scoring function
which integrates acoustic, syntactic, and semantic quality measures.

In the present paper we give a more detailed description of the
implementation of this approach combining probabilistic and symbolic
knowledge. First, we will illustrate why both contextual knowledge and
recognition scores are important for flexible robust parsing.
Next, the processing of these knowledge sources in the parser is
described.
Finally, we evaluate this approach by comparing the information content
rates of analysis results that were produced with and without semantic
predictions.

\section{CONTEXTUAL KNOWLEDGE AND RECOGNITION SCORE}
\label{CKRS}
The parser receives a word graph as its input. The nodes of the graph
represent points in time and the edges are labelled with scored word
hypotheses. 

\begin{figure}[htb]
\epsfxsize=7cm
\centerline{\epsffile{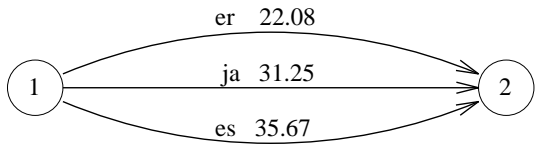}}
\label{WG1}
\caption{A simple word graph}
\end{figure}

Figure \ref{WG1} shows a simplified word graph that
contains three alternative one-word sentence hypotheses.
Scores are positive numbers which assign a (pseudo-)probability measure to
a word hypothesis: The smaller the score the
higher the probability, i.e., in the example graph the
hypothesis {\tt [1 er 22.08 2]} has the best score ({\tt 22.08}). 
If one adopts the traditional view
that it is the task of the parser to find the best scoring interpretation, then
we would expect the parser to deliver {\tt er} (he) as solution.

Now let us assume the following dialogue context:

\begin{tabular}{ll}
{\tt user1}: & {\em Ich m\"ochte morgen nach Ulm fahren.}\\
             & I want to go to Ulm tomorrow.\\
{\tt system1}: & {\em Sie wollen nach Ulm fahren?}\\
               & You want to go to Ulm?\\
\end{tabular}

Being a yes-no question, the last system utterance generates the
expectation that the interjections {\em ja} (yes) or {\em nein} (no) will
be contained in the user reply. Analyzing the graph in Figure \ref{WG1}
with this dialogue context, it is much more likely that the hypothesis {\tt [1 ja
  31.25 2]} is the correct solution, although it has not the best
score.  

Contextual expectations are mapped onto semantic predictions which are
passed down to the LP in our system (cf. Figure \ref{archi}). The
predictions are generated on the basis of the last system 
utterance in a way similar to the {\em dynamic prediction mechanism}
described in [1]. One possible 
way of using these predictions is to filter out all results
which are incompatible with the predictions. This strategy would have the
desired effect in the above example, but lead to a very restricted 
dialogue, because all additional user information were eliminated by 
this rigid filter. For example, in the context above a user might 
be overinformative and instead of simply confirming might answer: 

\begin{tabular}{ll}
{\tt user2}: & {\em Ja um zehn Uhr.}\\
             & Yes at ten o'clock.\\
\end{tabular}

Therefore, we prefer a less rigid strategy: If the semantic
content of a (partial) result agrees
with a top-down prediction, the result has a high pragmatic
relevance, otherwise low. Pragmatic relevance is expressed as a
numerical value which can be used to calculate a quality
score integrating scores of different processing levels. The
basic idea thus is to increase the overall score for predicted hypotheses in order
to compensate lower recognition scores.

\begin{figure}[htb]
\epsfysize=2cm
\epsfxsize=8cm
\centerline{\epsffile{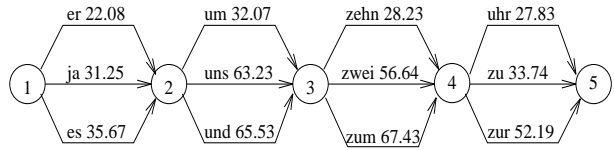}}
\label{graphex}
\caption{Word graph example}
\end{figure}

Assume, for example, that the word graph in Figure 3 was generated as the
recognizer output of analyzing utterance {\tt user2}.
In the context of a yes-no question we would like to increase the overall
score of the predicted hypothesis {\it ja} in the first part of the
utterance. But the overinformative second part of the utterance, {\em um
  zehn Uhr}, should still be acceptable as a parse result, although it may
not correspond to context expectations. The dialogue component of our
system is flexible enough to interpret such additional information (cf.~[4]).

\section{AN INTEGRATED QUALITY SCORE FOR CHART EDGES}

We use a chart-based island parser implemented in Prolog which looks for the best scored,
grammatically correct sentence hypothesis in the graph. It performs an
agenda-driven heuristic search (cf.~[5]). The chart of the parser is
initialized with the word  
hypotheses of the input graph. The linguistic knowledge base of the parser
is a highly lexicalized Unification Categorial Grammar (UCG) represented in
DATR (cf.~[2]). In UCG, syntactic and semantic structures are
represented and 
constructed in an integrated way. Example (\ref{Ja}) shows the lexical
entry of {\em ja}, which has the syntactic category {\it part} (particle)
and the semantic type {\it dm\_marker} (dialogue manager marker).
\begin{equation} 
\label{Ja}
\bra mor: \bra form:ja \ket \\[0.5ex]
     syn: \bra head: \bra major:part \ket \ket \\[0.5ex]
     sem: \bra type:dm\_marker, value:yes \ket
\ket
\end{equation}
Semantic predictions are provided from the
dialogue manager in a format compatible with the semantic representations
of lexical entries, e.g., the dialogue context ``yes-no question'' generates the
prediction list shown in (\ref{yes-no_preds}).
\begin{equation}
\label{yes-no_preds}
\bra
     \bra type:dm\_marker, value:yes \ket\\[1ex]
     \bra type:dm\_marker, value:no \ket\\[0.1ex]
\ket
\end{equation}
The semantic attribute-value pairs of both lexical entries and predictions
are compiled into Prolog terms with the same program.
Thus, agreement of a chart edge with semantic predictions can be checked
with standard Prolog unification. 

The predictions are used by the parser in two ways:
First, they serve as {\em seed definitions} for the island parser, which
can thus start its search from pragmatically relevant islands. Second,
the predictions contribute to the integrated quality
score which is assigned to each partial result during parsing.
In order to integrate the symbolic contextual knowledge with the numerical
recognition score we use a function $pr(E)$ which maps the agreement with a
prediction onto a numerical value.\footnote{A similar score called
  pragmatic priority was also used in the EVAR system (cf.~[7]).} In our
current experiments we use the 
following heuristic weightings: $pr(E)=4$ if the semantic type of a chart
edge $E$ unifies with one of the top-down predictions, otherwise $pr(E)=1$. 

The computation of the integrated quality score $QS$ of an edge $E$ is
defined as follows:
\begin{equation}
QS(E) = \frac{Q_a(E)}{sc(E) \times pr(E)} 
\end{equation} 
where $Q_a$ denotes the acoustic quality, $sc$ the value for syntactic
completeness\footnote{For the sake of simplicity we do not consider
  syntactic completeness in the following examples by setting $sc(E)=1$ 
  for all cases.}, and $pr$ the value for pragmatic relevance. The
interpretation of $QS$ is like that of the recognition score, i.e., the
smaller the better.

The acoustic quality $Q_a$ is given by: 
\begin{equation}
Q_a(E) = \frac{sf(E)}{length(E)} 
\end{equation}
The {\em shortfall} function $sf(E)$ (cf.~[9,~p.~298]) for a given edge
$E(i,j)$ that covers a segment from node $i$ to node $j$ is given by 
\begin{equation}
sf(E) = Maxseg - maxseg(i,j) + RS(E)
\end{equation}
where $Maxseg$ is the maximum total score of the whole graph, $maxseg(i,j)$ is
the maximum score of the segment $i$ to $j$, and $RS(E)$ is the recognition
score. 
For example, the {\em shortfall} of the hypothesis {\tt [1 ja 31.25 2]} in
Figure 3 is $110.21 - 22.08 + 31.25=119.38$, which reflects the
fact that a complete solution including this hypothesis is $9.17$ points worse
than the best scoring path {\tt [er um zehn uhr]}.

The $RS$ of an combined edge $CE$,
which was composed of two edges $E_1$ and $E_2$, is defined as the sum of
$E_1$ and $E_2$. 

Given these definitions, we can now illustrate the effect of semantic
predictions on parsing the word graph in Figure 3. Assume the graph
is parsed as an answer to a yes-no question, i.e., the prediction list
given in (\ref{yes-no_preds}) is used. Only one hypothesis, {\tt ja},
unifies with one of the predictions, {\tt
  [type:dm\_marker,value:yes]}, i.e., $pr(ja)=4$. Thus, its 
quality score is $\frac{119.38}{4}=29.84$, whereas the scores of the
alternative hypotheses spanning from node 1 to 2 stay
equal to the acoustic quality due to $pr(E)=1$.
Let us further assume that the grammar allows building a prepositional time
phrase {\tt um zehn uhr}. Since no time expression is predicted, the overall
quality score of this phrase is equal to the acoustic quality $Q_a$, i.e.,
$\frac{110.21 - 88.76 + 88.76}{3}= 36.74$.

Under the assumption that the grammar does not contain a rule to combine
{\tt ja} and {\tt um zehn uhr}, the parser will terminate without having found a complete
solution that spans the whole input. In this case, the robust mechanism of selecting
multiple partial results is applied: Starting from the edge with the best
quality score, the best scoring adjacent edges are collected recursively
until a sequence of partial results spanning the whole utterance is
found. In our example, the predicted result {\tt ja} has got the best
quality score during parsing. Since it is located at the beginning of the
graph, no left-adjacent solutions have to be looked for. Among the
right-adjacent edges {\tt um, uns, und, um zwei uhr} and {\tt um zehn uhr}
the latter has the best score and is 
selected. Its end node marks the end of the graph, too. Thus a sequence of
partial results through the graph was found and the LP hands over these
results as Semantic Interface Language (SIL, cf.~[6])
structures to the dialogue manager (cf. Figure \ref{archi}). Examples
(\ref{ParseRes1}) and (\ref{ParseRes2}) show the selected parsing 
results in SIL format.
\begin{equation}
\label{ParseRes1}
 \bra id:A\\
          syn: \bra id:B\\
                       category: part \\
                       string:ja\\
                       score:29.84
                  \ket \\[0.5ex]
          sem: \bra id:B\\
                          type:dm\_marker\\
                          value:yes
                     \ket
 \ket
\end{equation}
\begin{equation}
\label{ParseRes2}
\bra id:C\\
     syn: \bra id:D\\
                  category: prep \\
                  string:um\_ zehn\_uhr\\
                  score:36.74
              \ket \\[0.5ex]
      sem: \bra id:D\\
                       type:time\\
                       thehour: \bra id:E\\
                                          type:hour\\
                                          value:10
                                     \ket
                  \ket
\ket
\end{equation}

\section{EVALUATION}
The main task of a parser in a speech {\bf understanding} system is to
determine the meaning of the spoken utterance. It has been argued in
[8] that the sentence understanding capabilities of a SDS are best judged
by the information content (IC) metric. IC calculates the percentage of
task-relevant information (TRI) contained in the parser output. This
requires the annotation of each utterance with a series of attribute-value  
pairs, where each attribute is a task-relevant concept (TRC). In the
present domain of timetable inquiries, examples of TRCs are: {\tt
  source\-city, goal\-city, time, date, dm\_marker}. 
For example, the TRI of the utterance {\it ja um 10 Uhr} is
{\tt [dm\_marker:yes, time:10]}. This reference annotation is called
RTRI.

IC can then be calculated by comparing RTRI with the
parser output. For that purpose the SIL structures produced by the parser
are translated into attribute-value pairs compatible with the ones of
RTRI, e.g., the structure shown in (\ref{ParseRes2}) is mapped to
{\tt [time:10]}. The output of this translation is called PTRI.

Performance of the robust parser is measured by the metric
$IC$, which is calculated as a percentage using formula (\ref{ICform}) 
\begin{equation}
\label{ICform}
IC = 100 \left( 1-\frac{i+s+d}{items} \right)
\end{equation}
where $items$ is the total number of items in RTRI and $i$, $s$, and
$d$ are the numbers of items inserted, substituted, and deleted
in PTRI, respectively.\footnote{See [8] for a definition of how
  to calculate the number of insertions, deletions, and substitutions.}

Assume, for example, that the word graph in Figure 3, whose RTRI is {\tt
  [dm\_marker:yes, time:10]}, is parsed without predictions.
This will produce two partial results, namely {\tt er} and {\tt um 10 uhr}. The
SIL structure of the former cannot be mapped to a TRC. Thus PTRI is {\tt
 [time:10]}, i.e., $d=1$ because one of the RTRI items is deleted in
PTRI. This yields an IC of $100 \left(1-\frac{1}{2}\right)=50\%$.

The parser was tested in stand-alone mode on 50 word graphs generated by
the Daimler-Benz word recognizer [3]. The graphs had a density of 4 edges
per spoken word and a word accuracy rate of 73.3\%.

To measure the impact of the predictions we first parsed the graphs without
predictions. In the second setup, semantic predictions were handed over as
an additional argument to the parser. The choice of the prediction was
determined by the original dialogue context of the utterance.

The results are shown in the following table, where ic-pr and ic+pr are the
IC rates without and with predictions, respectively, and t-pr and t+pr are
the corresponding average parse times (in seconds) taken on a SPARCstation 10.
\begin{center}
\begin{tabular}{|r|r||r|r|} \hline
ic-pr & t-pr & ic+pr & t+pr\\ \hline
67.48 & 0.67 & 76.48 & 0.74\\ \hline
\end{tabular}
\end{center}
These figures show a 9\% increase of the IC rate when using
contextual knowledge in the parser. Most of this improvement was attained
in the analysis of very short, elliptical utterances typically provided as
answers to yes-no questions. In these cases, exemplified in Figure 2, the
linguistic grammar cannot contribute much so that merely contextual
expectations allow a well-founded choice among competing alternatives.

\section{CONCLUSION AND FURTHER WORK}
We have presented a mechanism for integrating contextual knowledge into the
linguistic processor of a spoken dialogue system. The reported results show
that the use of predictions can improve the IC rate
of the parser. Most improvement is gained in the analysis of
short utterances. To determine the IC, the
parsing results had to be inspected manually. In the near future an
annotated test suite of word graphs will be built up in order
to automate the evaluation.

Furthermore, we intend to measure the IC given different
processing time limits. As mentioned in section 3, the predictions are
used as pragmatic seed definitions of the island parser. Thus
partial results which are most relevant for understanding are
built at an early stage of processing. Due to this strategy we expect
acceptable IC rates even with a strong limit on processing
time as it may be necessary in a real time system.

\section{ACKNOWLEDGEMENTS}
This research was funded by the Daimler-Benz Research Center in Ulm. We
would like to thank Paul Heisterkamp, Wieland Eckert and
Christian Lieske.

\section{REFERENCES}

\parindent0pt
\parskip0.4ex
\baselineskip2.5ex
[1] F. Andry (1992). Static and Dynamic Predictions: A Method to Improve
Speech Understanding in Cooperative Dialogues. In {\em Proceedings of ICSLP
  92}, pp 639--642.

[2] F. Andry et.~al. (1992). Making DATR Work for Speech: Lexicon
Compilation in SUNDIAL. In {\em Computational Linguistics}, 18(3), pp
245--267.

[3] A. Brietzmann et.~al. (1994). Robust Speech Understanding.
 In {\em Proceedings of ICSLP 94}, pp 967--970.

[4] W. Eckert, H. Niemann (1994). Semantic Analysis in a Robust Spoken
Dialog System. In {\em Proceedings of ICSLP 94}, pp 107--110.

[5] G. Hanrieder, P. Heisterkamp (1994). Robust Analysis and Interpretation in
Speech Dialogue. In H. Niemann, R. De~Mori, G. Hanrieder (eds): {\em
  Progress and Prospects of Speech Research and Technology.} Infix: Sankt
Augustin, pp 204--211. 

[6] S. McGlashan, F. Andry, and G. Niedermair (1990). A Proposal for
SIL. ESPRIT Projekt P2218 SUNDIAL, technical report.

[7] H. Niemann et.~al. (1988). A Knowledge Ba\-sed Speech
Understanding System. In {\em International Journal of Pattern
  Recognition}, 2(2), pp 321--350.

[8] A. Simpson, N. Fraser (1993). Black Box and Glass Box Evaluation of the
SUNDIAL System. In {\em Proceedings of EUROSPEECH 93}, pp 1423--1426.

[9] W. Woods (1982). Optimal Search Strategies for Speech Understanding
Control. In: {\em Artificial Intelligence}, 18, pp 295--326.
\end{document}